\documentclass{article}

\usepackage[final]{neurips_2024}

\usepackage{amsmath}
\usepackage{amsfonts}
\usepackage{amssymb}
\usepackage{graphicx}
\usepackage{hyperref}

\usepackage{algorithm}
\usepackage{algorithmic}

\usepackage{natbib}
\usepackage{booktabs}

\title{EmoAra: Emotion-Preserving English Speech Transcription and Cross-Lingual Translation with Arabic Text-to-Speech}

\author{
Besher Hassan
\And
Ibrahim Alsarraj
\And
Musaab Hasan
\And
Yousef Melhim 
\And
Shahem Fadi
\And
Shahem Sultan
\AND
\normalfont
Email: \small \texttt{Besher.Hassan@mbzuai.ac.ae}, \texttt{Ibrahim.Alsarraj@mbzuai.ac.ae}
 \\
}

\begin{document}

\maketitle

\begin{abstract}

This project addresses challenges in cross-linguistic communication, particularly in banking customer service, where understanding emotional context is essential. By addressing language barriers, fostering empathetic communication, and enhancing client satisfaction, the system demonstrates practical applications in banking customer service. It integrates \textbf{Speech Emotion Recognition (SER)}, \textbf{Automatic Speech Recognition (ASR)}, \textbf{Machine Translation (MT)}, and \textbf{Text-to-Speech (TTS)} to enable emotion-preserving, multilingual communication. Using a \textbf{CNN} for emotion detection, \textbf{Whisper} for English transcription, \textbf{MarianMT} for English-to-Arabic translation, and \textbf{MMS-TTS-Ara} for Arabic speech synthesis, the system processes spoken English, detects emotions, transcribes and translates the text, and delivers a synthesized Arabic response while retaining emotional nuance. Key findings include an F1-score of \textbf{94\%} for emotion detection, a BLEU score of \textbf{56}, a BERTScore F1 of \textbf{88\%} for translation quality, and \textbf{81\%} in human evaluation for translation accuracy, fluency, and domain-specific terminology. For further details, check the \href{https://github.com/besherhasan/Emotion-Driven-Speech-Transcription-and-Cross-Lingual-Translation-with-Arabic-TTS-Integration}{GitHub Repository}.

\end{abstract}

\section{Introduction}
\subsection{Project Overview}
In an increasingly globalized world, businesses face the challenge of communicating effectively across language and cultural barriers. The goal of this project is to build an emotion recognition system for applications in customer service specifically within the banking sector. By extracting the emotion from a customer’s spoken interaction, and then transcribe the English audio, translate the text into Arabic, and use the text-to-speech (TTS) system for oral presentation. Combine Emergent Speech Emotion Recognition (emergent SER), Automatic Speech Recognition (ASR), Machine Translating (MT) with text-to-speech (TTS) and create multilingual communication, retaining emotional intactness. The project will be language-agnostic and emotion-oriented. \textit{Emotion-oriented systems} are systems designed to recognize and process content to understand and preserve the emotional context in communication.
In customer service contexts, it is common to look at the emotional tone in a customer's voice which helps in gauge what the customer needs and feels in particular. The importance gets magnified in case of banking since emotions help in Assessing the degree of urgency, satisfaction, or dissatisfaction. Unfortunately, language barriers and lack of live emotional responses act as a hindrance in enabling the bank representatives from properly addressing the issues on time. Our project utilizes emotion and language understanding to ease these barriers and give the industry the ability to respond emotionally to emotionally charged speech in any language.

\subsection{{Primary Contributions}}
 the primary contributions of this project can be summarized as:

    \begin{itemize}
    \item Emotion in Cross-Linguistic Interactions: In contrast to many existing systems which ignore emotional tone in translations, translation done in this project has an emotional concern as to the fact that emotions in the original language will definitely be present in the target language as well, making clients feel more natural and more empathy in the service provision.
    
    
    \item Practical Application in Banking Customer Service: A typical application of our system falls within the banking and finance industry where communication across different languages for effective service delivery is possible as well as the need to understand customers emotions.

    \end{itemize}

\subsection{Related Work}
Doctors Livingstone and Russo (2018) provided the Ryerson Audio-Visual Database of Emotional Speech and Song (RAVDESS), which has now become a widely used benchmark for assessment of SER systems \citep{livingstone2018ryerson}. Employing this dataset, researchers have used CNNs and RNNs for emotion recognition for a wider variety of emotions.

Huang et al. (2021) showed how models for emotion detection can be fine-tuned through the transfer learning process towards achieving better accuracy by retraining existing models \citep{huang2021speech}.

Hannun et al. (2014) introduced the DeepSpeech system which leveraged recurrent neural networks trained on massive datasets of speech and achieved impressive results \citep{hannun2014deepspeech}. Lately, Baevski et al. (2020) presented wav2vec 2.0 which enables the self-training of networks which produce better speech representations than what has been previously achieved using end-to-end architectures \citep{baevski2020wav2vec}.

Machine Translation (MT) has made one of the most rapid advancements as well with special emphasis on specific low-overhead language models. Junczys-Dowmunt et al. (2018) introduced MarianMT system which is a quick, powerful, and elegant neural machine translation system and achieved the best results in neural translation for several language pairs \citep{junczys-dowmunt2018marian}.

Text-to-Speech (TTS) systems have developed over the years with the help of technologies such as Google Cloud’s TTS service which produces nearly natural speech. Several languages including Arabic are supported and systems like these have been used in a number of practical cases. But still, the majority of existing TTS systems are designed in such a way that their functions are separated from systems of emotion recognition and machine translation. This, in turn, renders these systems less effective for dynamic and emotionally complex applications \citep{google2021tts}.

Building on these ideas and undertaking the relevant research work, this project seeks integration of ASR, MT, SER, and TTS methods into a single, coherent system. In particular, we want to create a powerful end-to-end multilingual system that is capable of preserving the emotionality of the content during the language conversion pipeline through the application of transfer-learning methods and pretrained models.

\section{Problem Statement}
In the present interconnected world, it is crucial that people who offer a service and the customers know how to effectively communicate across different languages particularly in fields like banking where the customer’s emotion can be valuable in determining the service offered. Unfortunately, most automated customer service systems are unable to detect and interpret emotions embedded in speech in the first and different languages. Such a impediment may contribute to miscommunication, less satisfaction of the customer and failure to engage in responsive communication. Our objective is to address this gap by developing a system that not only transcribes and translates spoken language but also retains the emotional context within customer interactions. By implementing a multilingual, emotion-driven speech recognition and translation pipeline, our solution will empower customer service teams to respond more empathetically and accurately to customers’ needs, regardless of language differences.

\section{Proposed Method}

\subsection{Exploratory Data Analysis}

\subsubsection{Translation Data Preprocessing}

To prepare the translation datasets for fine-tuning the Marian MT model, preprocessing steps were applied to make the data compatible with the model:

\begin{itemize}
 \item \textbf{Text Normalization}: Lowering case, deleting numbers and punctuation, and cutting excessive text space. 
 
    \item \textbf{Language-Specific Filtering}: 
To warrant consistency of the data submitted, only the Arabic script was kept on the Arabic data and English alphabetic letters only in English data.
 
 \item \textbf{Tokenization and Padding}: Sentences were tokenized using MarianMT tokenizer, then truncating sequences to a maximum length of 128 tokens to achieve a balance between memory efficient and text coverage.

\end{itemize}
These preprocessing steps prepared the datasets for fine-tuning by standardizing the content, maintaining linguistic consistency, and reducing errors during training.

\subsubsection{Audio Files}
\begin{figure}[htbp]
\centering
\includegraphics[width=0.98\textwidth]{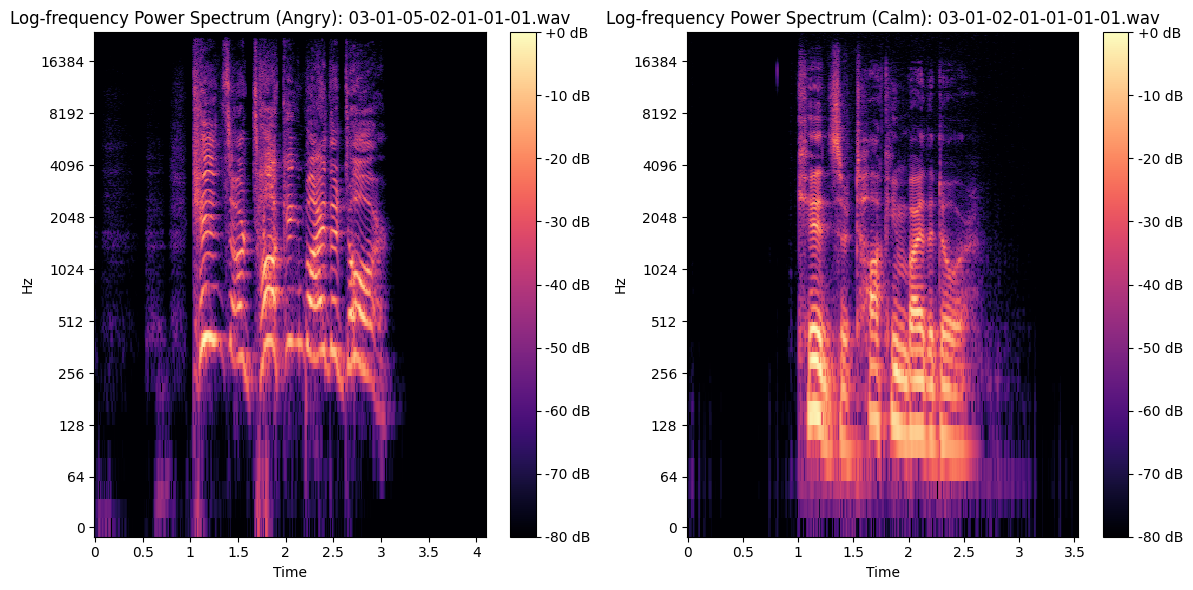} 
\caption{Log-Frequency Power Spectrum of Different Emotions}
\label{fig:spectrograms}
\end{figure}

There are 1440 audio files in .wav format. The audio files tagged with `angry` and `calm` have distinctly different frequency and amplitude characteristics as seen in Figure \ref{fig:spectrograms}. The left spectrogram illustrates the log-frequency power spectrum of an "Angry" speech sample, characterized by higher intensity and wider frequency variations. The right spectrogram represents the "Calm" emotion, showing lower intensity and smoother frequency transitions. These differences highlight the distinct acoustic features of emotional speech, particularly in terms of energy distribution across frequencies and time. 

\subsection{Method}
We built an end-to-end system that consists of four components as shown in Figure \ref{fig:System Arch}, \textbf{CNN} model for extracting the emotion. (\textit{LSTM and ResNet50 models were implemented and tested instead of CNN, more details can be found in} \ref{sec:AddRes}). Followed by \textbf{Whisper}  \cite{openai_whisper} pre-trained transformer model to recognize the speech and convert it to text. Then the English sentences are translated into Arabic sentences using \textbf{MarianMT} \cite{huggingface_marian} pre-trained fine-tuned transformer model. Finally, the text will be converted to speech using a \textbf{MMS-TTS-Ara} \cite{facebook_mms_tts_ara} pre-trained model.
Their respective roles in the processing pipeline:

\begin{figure}[b]
\centering
\includegraphics[width=0.98\textwidth]{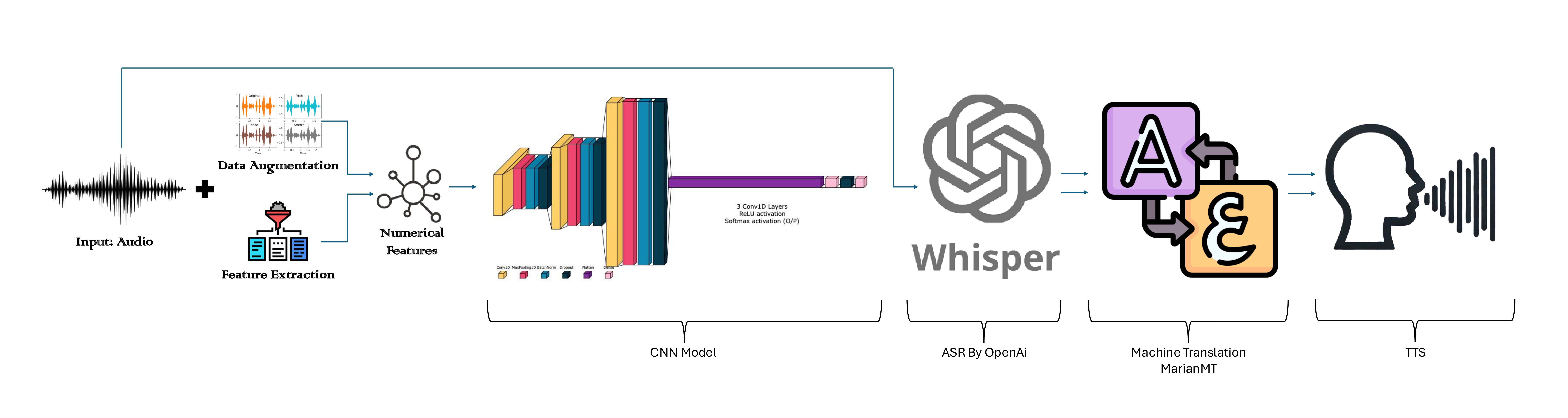} 
\caption{System Architecture}
\label{fig:System Arch}
\end{figure}

\subsubsection{CNN Model for Speech Emotion Recognition:}

This component employs an artificial neural network model that is created in TensorFlow/Keras to recognize emotion in audio data. The following was implemented:

\textbf{Data Preparation:}
In order have better generalization, numerical features from (emotion-features.csv) was created and \textbf{data augmentation} techniques were applied to the audio files to generate multiple variations of the same audio file to simulate different acoustic conditions. Below are the applied techniques:

\begin{itemize}
    \item \textbf{Adding Noise:}
    Gaussian noise is added to the audio signal.
\begin{equation}
y_{\text{noisy}} = y + \text{noise\_factor} \cdot \mathcal{N}(0,1)
\end{equation}

\text{where } \( y \) \text{ is the original signal and } \( \mathcal{N}(0, 1) \) \text{ is a standard normal distribution.}

    \item \textbf{Pitch Shifting:}
     Adjusts the pitch of the signal by shifting it up or down in frequency.

     \item \textbf{Time Stretching:}
     Speeds up or slows down the audio signal.

     \item \textbf{Time Shifting:}
     Shifts the signal by a percentage of its length.

\end{itemize}

Following the data augmentation, \textbf{feature extraction} was used to convert the audio waveforms into numerical representations, which is a dimensionality reduction step for keeping only the relevant information of the data. Several feature extraction approaches were used:

\begin{itemize}
    \item \textbf{Zero-Crossing Rate (ZCR)}
     By measuring the rate at which the signal crosses the zero amplitude line, capturing the noisiness of the signal.
     \begin{equation}
\text{ZCR} = \frac{1}{T} \sum_{t=1}^{T} 1\left[ y_t \cdot y_{t-1} < 0 \right]
\end{equation}

\text{where } T \text{ is the number of samples, and } 1 \text{ is an indicator function.}

    \item \textbf{Root Mean Square Energy (RMSE)}
    Measures the energy of the audio signal.
    \begin{equation}
\text{RMSE} = \sqrt{\frac{1}{T} \sum_{t=1}^{T} y_t^2}
\end{equation}

    \item \textbf{Mel-Frequency Cepstral Coefficients (MFCCs)}
    Captures the spectral properties of the signal and represents its timbre.

    \begin{equation}
\text{MFCC}[k] = \sum_{n=0}^{N-1} \log(X[n]) \cdot \cos\left( \frac{N \pi k}{N} (n - 0.5) \right)
\end{equation}

\text{where } X[n] \text{ is the Mel-scaled power spectrum.}

\end{itemize}

\label{sec:Setup}

After the feature extraction, the data will be ready to be fed into the CNN model. The construction of CNN is as the following:

\textbf{One-Hot Encoding:}
Emotion labels (Y) were categorical and were one-hot encoded using OneHotEncoder. This converts the labels into binary vectors which are suitable for multi-class classification.

\textbf{Feature Scaling:}
All features were normalized using the (StandardScaler), it scales the data to have zero mean and unit variance to ensure the numerical stability during optimization.

\textbf{Reshaping for CNN Input:}
The features got reshaped to (samples,features,1), making them compatible with the 1D convolutional layers.

The CNN model was designed with the following layers:

\textbf{Input Layer:}
Accepts input data of shape (features,1) which represents the scaled features.

\textbf{Convolutional Layers:} The audio signals undergo three 1D convolutional layers (with filters) that extract features with ReLU activations to non-linearities that are used to address vanishing gradients.

\textbf{Batch Normalization:}
Normalizes the outputs of each layer to accelerate training.

\textbf{Max Pooling and Dropout Layers:} Each of the convolution layers has max pooling as a subsequent layer in a bid to reduce the feature dimensionalities, and dropout for convectional layers and for dense layers to prevent overfitting.

\textbf{Fully Connected Dense Layers:} After the convolutional output has been flattened, a dense layer of 256 neurons has been used with ReLU activation and the softmax output is done in multi-class after classifying the images.

The model is compiled using the Adam optimizer, which is known for its efficiency in training deep neural networks.
Categorical cross-entropy is used as the loss function, as the problem is a multi-class classification task. Overall, CNNs are effective for capturing patterns in sequential data. The use of 1D convolution layers matches the structure of the audio features.

\subsubsection{Whisper by OpenAI}
Whisper is an advanced automatic speech recognition (ASR) system developed by OpenAI \cite{openai_whisper}. Its architecture is based on a transformer-based model. Key components:

\textbf{Encoder-Decoder Architecture:}
Whisper adopts an encoder-decoder approach where in this case the encoder is responsible for the raw audio while the decoder generates the transcription from the encoded features. Using this sequence-to-sequence framework, the decoder takes in the speech signal and produces text tokens.

For our project, we are using \textbf{Whisper Base} \cite{openai_whisper_github}. Different Whisper variants offering speed and accuracy trade-offs. In general, Whisper is very robust against variations like accents, noise, and background disturbances which makes it a reliable ASR model.

\subsubsection{English To Arabic Translation}

In this section, the focus is to enhance English-to-Arabic translations in the banking field through the use of a fine-tuned Marian MT. First, attempts to provide translation through multilingual models were made. Specific steps that included evaluation of pre-training, preparation of training dataset, fine-tuning and, evaluation metrics incorporating various metrics were utilized. Domain-specific data and advanced transformer-based architecture has been use.

\textbf{Dataset Preparation}

The dataset included a custom created banking translation dataset and an English to Arabic general translation dataset. These were divided into training, validation, and testing subsets with the proportions of 80\%, 10\% and 10\% respectively. 

\textbf{Preliminary Model and Challenges}

The objective of this work was to develop a high-performing translation model. Due to the limited dataset size, training a custom-built Transformer model posed significant challenges. Initial efforts involved training a Transformer model on 24,000 English–Arabic sentence pairs. This approach provided basic translation capabilities but was constrained by the dataset's size, which was insufficient for effective generalization—particularly for complex language pairs like English and Arabic. Building a high-performing Transformer model typically requires millions of sentence pairs to achieve the desired results. To address these constraints, we fine-tuned a pre-trained model, MarianMT, to optimize translation performance on the available small-scale dataset.

\textbf{Adoption of Marian MT}

Marian MT, a pre-trained translation model by Helsinki-NLP \cite{helsinki-nlp}, is a transfomer-sequence-model highly efficient based. Another way to consider it is that it consists of an encoder, a decoder and multi-head self attention mechanism and layer normalization and position-wise feed forward networks. The source language text about contextual and syntactic structures is captured by the encoder as a series of continuous representations. Then, the encoder transforms the source text into several continuous representations. The decoder then generates a target text by attending to the representations and modeling the dependencies in a target sequence. This has emphasized the importance of pre-trained models and extra data this made MariaMT as a suitable choice.

\textbf{Hyperparameter Experimentation}

In order to undertake training for the Marian MT model optimization, experiments with some key hyperparameters were carried out, such as learning rate, batch size, number of epochs, beam search width, and gradient accumulation steps. and  several configurations were tested to find the optimal hyperparameters values that gives the best translation quality.

\textbf{Mathematical Formulation}

The model minimizes the cross-entropy loss \( \mathcal{L} \) between the predicted tokens \( \hat{y} \) and the reference tokens \( y \), defined as:
\[
\mathcal{L} = -\frac{1}{N} \sum_{i=1}^{N} \log P(\hat{y}_i | y_{<i}, x)
\]
where \( x \) is the input sequence, \( y_{<i} \) are preceding tokens, and \( N \) is the number of tokens.

MarianMT efficiency and being pre-trained on different language datasets provided a strong baseline. In addition, the ability of transformer architecture to model long range dependencies is essential in performing complex translations such as English-Arabic.

\subsubsection{Text-to-Speech Using MMS-TTS-Ara}

The MMS-TTS-Ara pre trained model is utilized in the final stage of the pipeline as shown in Figure \ref{fig:System Arch}, to convert translated Arabic text into natural, fluent speech. It integrates a \textbf{text encoder}, \textbf{sequence generator}, and \textbf{neural vocoder} to produce high-quality audio output.

\begin{itemize}
    \item \textbf{Text Encoder:} Converts Arabic text into latent representations, capturing phonetic and syntactic details using a Transformer-based architecture.
    \item \textbf{Sequence Generator:} Maps these representations to mel-spectrograms, encoding the time-frequency characteristics of speech.
    \item \textbf{Neural Vocoder:} Transforms mel-spectrograms into waveform audio using advanced models like HiFi-GAN, producing human-like speech with clear intonation.
\end{itemize}

The process involves text normalization, tokenization, and adjustments to prosody, ensuring the retention of emotional context from upstream components. MMS-TTS-Ara’s optimization for Arabic guarantees fluent, natural output, making it ideal for banking customer interactions. This step completes the multilingual communication pipeline with a seamless transition from text to speech.

\section{Experimental Setup}

\subsection{MarianMT Fine-Tuning}

\begin{figure}[t]
    \centering
    \includegraphics[width=0.9\textwidth]{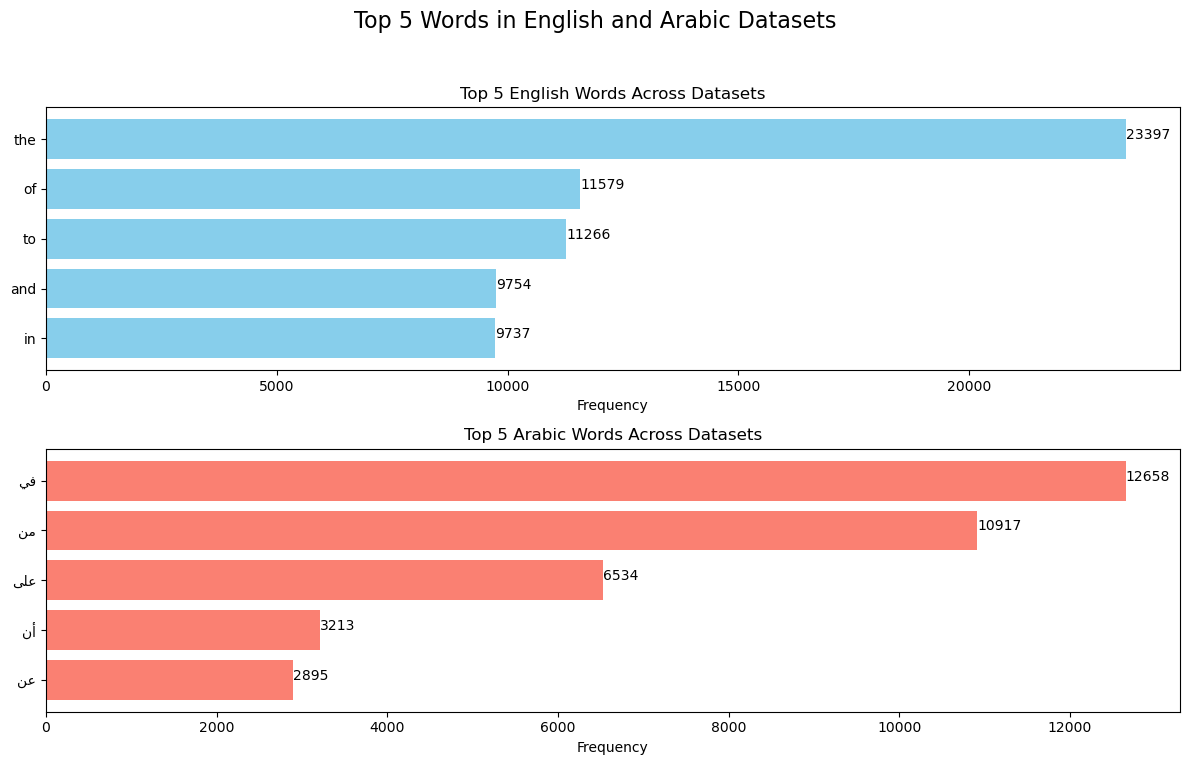}
    \caption{Top 5 Words in English and Arabic Dataset}
    \label{fig:top5}
\end{figure}

In order to integrate the Marian MT model within the banking domain, a methodological fine-tuning process was implemented, aiming to augment the translation quality while maximizing training efficiency. The translation datasets employed for fine-tuning show a high occurrence of function words, such as “the,” “of,” and “to” in English, as well as similar words in Arabic, as shown in Figure \ref{fig:top5}. Such overused words are essential to help the model comprehend the most basic structural rules of any language and its grammar, making it easier to translate accurately and fluently. In terms of sentence complexity, the Arabic and English banking datasets comprised an average sentence length of 10 words and 11 words respectively, whereas the general English-to-Arabic dataset had longer averages of 15 in Arabic and 17 in English. This range in sentence length likely helped the model’s flexibility, allowing it to work efficiently with both short and complex sentences.

\subsection{Datasets}

\subsubsection{RAVDESS Dataset}
 In particular, RAVDESS \citep{ravdess2023} dataset is shown in the table \ref{table:datasets}, will be used for the task of training our Speech Emotion Recognition (SER) model. each filename has identifiers as per the official RAVDESS website: (XX-XX-XX-XX-XX-XX-XX.wav)

Modality (01 = full-AV, 02 = video-only, 03 = audio-only).
Vocal channel (01 = speech, 02 = song).
Emotion (01 = neutral, 02 = calm, 03 = happy, 04 = sad, 05 = angry, 06 = fearful, 07 = disgust, 08 = surprised).
Emotional intensity (01 = normal, 02 = strong). NOTE: There is no strong intensity for the 'neutral' emotion.
Statement (01 = "Kids are talking by the door", 02 = "Dogs are sitting by the door").
Repetition (01 = 1st repetition, 02 = 2nd repetition).
Actor (01 to 24. Odd numbered actors are male, even numbered actors are female). The dataset was split into training, validation, and test sets. The splitting is 7.5\% for testing, 7.5\% for validation, and 85\% for training.

\begin{table}[htbp]
\centering
\begin{tabular}{|p{0.48\textwidth}|p{0.48\textwidth}|}
\hline
\textbf{Dataset} & \textbf{RAVDESS} \\ 
\hline
\textbf{Modalities} & Audio, Video (Full-AV), Audio-only \\
\hline
\textbf{Language} & English (North American accent) \\
\hline
\textbf{Number of Samples} & 1440 audio files \\
\hline
\textbf{Number of Participants} & 24 actors (12 male, 12 female) \\
\hline
\textbf{Number of Emotion Classes} & 8 \\
\hline
\end{tabular}
\caption{Characteristics of the RAVDESS dataset}
\label{table:datasets}
\end{table}

\subsubsection{Arabic to English Translation Dataset}

For the translation part we used two datasets in order to effectively finetune the MarianMT model, the first one is English to arabic Translation sentences \cite{moustafa2024arabictoenglish} from Kaggle that centres around general topics in MSA, The dataset is sized at 6.4 MB and it has in excess of 24000 samples as depicted in english and its translation to arabic.

\subsubsection{Banking 77 Dataset}
For the second part of the dataset we have created a new dataset from the \textbf{Banking 77} \cite{banking77_huggingface} Dataset which is available on Hugging Face's dataset repository, and this dataset originally was in english language so we used Google Translator and MyMemoryTranslator models from the deep translator to translate the sentences to Arabic and after that we have verified that the translation of the translated sentences is correct. the data set contain 10000 sentences in english and their respective translation in Arabic, and the size of the dataset is 773 KB.

\subsection{Hyper-parameters}
Different hyper-parameters were tested for training the CNN model. Moreover, different hyper-parameters configurations were tested as well for the translation model to identify the optimal setup for fine-tuning. Below are the key approaches used:

\begin{itemize}
    \item \textbf{High Quality:} Focused on accuracy and fluency with a high beam search width, lower learning rates, and longer epochs.
    \item \textbf{Balanced:} Used a moderate learning rate, fewer epochs, and smaller beam searches for effective training within a reasonable timeframe.
    \item \textbf{Fast Training:} Adopted higher learning rates, fewer epochs, and smaller batch sizes to complete training quickly at the expense of quality.
\end{itemize}

After testing these configurations, the optimal hyperparameters for fine-tuning are detailed in Table ~\ref{tab:hyperparameters} under the section "Fine-tuned MarianMT Model Hyperparameters."

\begin{table}[ht]
\centering
\begin{tabular}{|p{0.30\textwidth}|p{0.66\textwidth}|}
\hline
\textbf{Hyperparameter} & \textbf{Value} \\
\hline
\multicolumn{2}{|c|}{\textbf{CNN Model Hyperparameters}} \\
\hline
Learning Rate & $1 \times 10^{-3}$ \\
Epochs & 100 \\
Batch Size & 64 \\
Kernel Size & 3 \\
Dropout Rates & 0.15 (Conv layers), 0.25 (Dense layer) \\
Number of Filters & 64, 128, 256 \\
Pool Size & 2 \\
Optimizer & Adam \\
Activation Function & ReLU (Hidden layers), Softmax (Output layer) \\
\hline
\multicolumn{2}{|c|}{\textbf{Fine-tuned MarianMT Model Hyperparameters}} \\
\hline
Batch Size & Per-device batch size of 8 with gradient accumulation steps of 4, resulting in an effective batch size of 32. \\
Learning Rate & $3 \times 10^{-5}$ with 1000 warmup steps for stable training. \\
Epochs & Fine-tuned over 10 epochs for optimal convergence. \\
Generation Parameters & Maximum sequence length of 300 tokens and beam search width of 8. \\
Weight Decay & 0.01 for regularization. \\
Evaluation Strategy & Performed at the end of each epoch. \\
Mixed Precision & FP16 enabled for faster training and reduced memory usage. \\
Gradient Accumulation & Accumulation over 4 steps for memory efficiency. \\
Save Strategy & Best model saved at the end of each epoch, limited to 2 checkpoints. \\
Optimization & Adam optimizer with weight decay for stability. \\
Validation Batch Size & Per-device evaluation batch size of 8. \\
\hline
\end{tabular}
\caption{Hyperparameters for CNN and fine-tuned MarianMT model.}
\label{tab:hyperparameters}
\end{table}

\subsection{Evaluation Metrics}

\subsubsection{Emotion Classification}
For evaluating the Speech Emotion Recognition (SER) component, different metrics might be used, like: precision, recall, F1-score, and support.

\subsubsection{Translation}
\label{sec:translation}
To evaluate the translation quality a combination of BLEU score(Translation quality by measuring n-gram between model outputs and reference) BERTScore(Compares contextual embeddings to evaluate meaning and fluency). since the automated metrics have limitation to give the correct score, a Human Evaluation was implemented on 100 test samples from banking-domain translations dataset. Each translation was rated of a scale from 1 to 3, with 0.5-point increments, based on accuracy, fluency, and domain-specific terminology As shown in Table~\ref{table:translation_quality}:
\begin{table}[htbp]
\centering
\begin{tabular}{|p{0.18\textwidth}|p{0.78\textwidth}|}
\hline
\textbf{Score} & \textbf{Description} \\ \hline
\textbf{3}     & Highly accurate, fluent, and domain-specific, resembling human translation. \\ \hline
\textbf{2.5}   & Mostly accurate with minor phrasing or terminology inconsistencies. \\ \hline
\textbf{2}     & Acceptable, conveying the main meaning with some minor inaccuracies. \\ \hline
\textbf{1.5}   & Noticeable issues in accuracy or fluency, causing slight confusion. \\ \hline
\textbf{1}     & Significant issues, making the translation difficult to understand. \\ \hline
\end{tabular}
\caption{Translation quality evaluation criteria}
\label{table:translation_quality}
\end{table}

The human evaluation gave a better insights about the quality if the translation on the banking domain, addressing nuances missed by automated metrics and enhancing the overall assessment.

\subsection{Baseline Comparison}

Baseline models were created to assess the performance improvements achieved through fine-tuning and additional processing. These baselines provide a reference for evaluating the effectiveness of the developed system:

\begin{itemize}
    \item \textbf{CNN Model:} A baseline CNN model was created without fine-tuning, data augmentation, or feature extraction. 
    
    \item \textbf{MarianMT Model:} For translation tasks, the MarianMT model in its pre-trained state was used as a baseline. Its performance was compared to the fine-tuned version to evaluate the improvements in translation quality.
\end{itemize}

These baselines establish a foundation for assessing the impact of enhancements introduced during model development.

\subsection{Computational Sources}
The project runs on RTX 3060 GPU, Ryzen 3700X CPU, and 64GB RAM, while Google Colab GPUs like: A100 and T4 will also be employed for supporting computational resources.

\section{Results \& Discussion}
\subsection{CNN Model}
\label{sec:CNNxRes}

\begin{table}[htbp]
\centering
\begin{tabular}{p{0.34\textwidth} p{0.30\textwidth} p{0.30\textwidth}}
\toprule
\textbf{Emotion Class} &
\textbf{Baseline CNN (F1-Score)} &
\textbf{Implemented CNN (F1-Score)} \\
\midrule
Angry       & 0.66 & 0.97 \\
Calm        & 0.49 & 0.92 \\
Disgust     & 0.52 & 0.96 \\
Fearful     & 0.48 & 0.92 \\
Happy       & 0.56 & 0.94 \\
Neutral     & 0.52 & 0.93 \\
Sad         & 0.50 & 0.92 \\
Surprised   & 0.53 & 0.95 \\
\midrule
\textbf{Average F1-Score} & \textbf{0.53} & \textbf{0.94} \\
\bottomrule
\end{tabular}
\caption{F1-scores of baseline and implemented CNN models across emotion classes}
\label{tab:cnn_f1_scores}
\end{table}

The table presented above \ref{tab:cnn_f1_scores} shows the performance of the implemented CNN model and the Baseline CNN in F1-score. Results show that the implemented CNN model performed better than the baseline across all the classes of emotions. Calm, fearful and sad emotion classes achieved the lowest F1 score in both models, this indicates that both models are somehow facing difficulties classifying these emotions, possible reason might be due to their overlapping acoustic features. On the other hand, angry emotion class scored the highest F1 score in both models among all different emotion classes, this might be from the unique and high intensity power spectrum of angry emotion class shown previously \ref{fig:spectrograms}. More importantly, the results prove how essential are feature extraction, fine-tuning, and data augmentation for developing SER systems. The Baseline CNN did not use any of these features and thus found it difficult to classify the different states of emotions and so performed poorly across all the metrics. However, the implemented CNN was able to get good F1 score which makes it applicable in real life. This demonstrates the importance of leveraging specific features and robust training techniques while developing state-of-the-art models. \textit{Different models were tested instead of CNN, more details can be found in} \ref{sec:AddRes}. 

\subsection{Translation}
Training a Transformer model from scratch yields a BLEU score of 23.00, highlighting challenges that we might face with a small dataset; using a pre-trained Marian MT model improves this score slightly to 25.48. Fine-tuning on both domain-specific and general datasets leads to significant gains—the best model reaches a BLEU score of 56 and a BERTScore F1 of 88.7\% as shown in the Table \ref{table:scores}. Despite these improvements, it might encounter issues with very long sentences: the model might truncate translations or misinterpret context due to the training dataset, since it does not contain sentences of more than 20 words which constrain its ability to capture dependencies in lengthy sequences. Shorter sentences are handled more effectively. Integrating domain-specific and general datasets allows the model to learn banking-specific terminology while maintaining broader language capabilities, enabling the model to generalize across major sentence structures and contexts and achieve higher BLEU and BERTScore metrics. Human evaluation, discussed in Section ~\ref{sec:translation}, supports these findings: scoring 100 test samples from the banking domain on accuracy, fluency, and domain-specific terminology the criteria are detailed in ~\ref{table:translation_quality}, The model achieves an average human evaluation score of 81\%, aligning closely with automated metrics and confirming its capability in domain-specific translation tasks.

\begin{table}[htbp]
\centering
\begin{tabular}{|p{0.32\textwidth}|p{0.18\textwidth}|p{0.22\textwidth}|p{0.20\textwidth}|}
\hline
\textbf{Model} &
\textbf{BLEU Score} &
\textbf{BLEU CI} &
\textbf{BERTScore F1} \\
\hline
From-Scratch Transformer
& 23.78
& (22.94--24.5)
& 67\% \\
\hline
Pre-trained Marian MT
& 25.48
& (23.98--27.09)
& 73\% \\
\hline
\texttt{Fine-Tuned MarianMT Model}
& 56.00
& (54.5--57.5)
& 88.7\% \\
\hline
\end{tabular}
\caption{BLEU and BERTScore results with confidence intervals}
\label{table:scores}
\end{table}

The results highlight the power of fine-tuning for domain-specific translation, with improved BLEU, BERTScore, and human evaluation metrics. Combining domain-specific and general
datasets significantly enhanced translation quality in specialized fields like banking.

\section{Limitations}
The fine-tuned Marian MT model showed an enhanced translation quality but it was limited by the dataset size. increasing the number of samples in the banking-domain dataset could further enhance accuracy and fluency. Future work could explore data expansion through back translation or paraphrasing with models like T5 and advanced fine-tuning techniques such as adapter layers or LoRA. Additionally, we can assess the quality of the translation by taking sample of it and give to human to give their feed back on its quality which will give an accurate evaluation.

\section{Conclusion}

This project was able to successfully demonstrate the integration of emotion-driven speech transcription and cross-lingual translation within the banking customer service environment. By leveraging several models, including a CNN for speech emotion recognition, Whisper for automatic speech recognition, MarianMT for translation, and MMS-TTS-Ara for text-to-speech synthesis, we developed an end-to-end system capable of preserving emotional integrity across languages. The system's practical applicability in real-world banking scenarios underscores its ability to enhance customer experience by addressing language barriers and emotional nuances. Future work will focus on expanding datasets and exploring advanced fine-tuning techniques to further refine accuracy and fluency in translations.

\appendix

\section{Additional Results}
\label{sec:AddRes}

\section{LSTM \& ResNet50}
\begin{table}[htbp]
\centering
\begin{tabular}{p{0.32\textwidth} p{0.30\textwidth} p{0.30\textwidth}}
\toprule
\textbf{Emotion Class} &
\textbf{LSTM (F1-Score)} &
\textbf{ResNet50 (F1-Score)} \\
\midrule
Angry       & 0.85 & 0.90 \\
Calm        & 0.83 & 0.86 \\
Disgust     & 0.86 & 0.91 \\
Fearful     & 0.84 & 0.88 \\
Happy       & 0.88 & 0.92 \\
Neutral     & 0.83 & 0.87 \\
Sad         & 0.82 & 0.85 \\
Surprised   & 0.89 & 0.93 \\
\midrule
\textbf{Average F1-Score} & \textbf{0.87} & \textbf{0.89} \\
\bottomrule
\end{tabular}
\caption{F1-scores of LSTM and ResNet50 models across emotion classes}
\label{tab:lstm_resnet_f1_scores}
\end{table}

As shown in Table \ref{tab:lstm_resnet_f1_scores}, LSTM and ResNet50 are additional two models implemented and tested across all different emotion classes. Both models have almost the same average F1 score. However, ResNet50 scored a higher score in all emotion classes, this is probably due to the deep residual architecture of ResNet50 which allows it to extract robust features from the audio signals. As noticed earlier for sad, fearful and calm emotion classes in section \ref{sec:CNNxRes}, both LSTM and ResNet50 models have the same difficulties differentiating different emotion classes.

\bibliographystyle{plain}
\bibliography{references}

\end{document}